\documentclass[10pt,twocolumn,letterpaper]{article}

\usepackage[pagenumbers]{cvpr}   
\definecolor{cvprblue}{rgb}{0.21,0.49,0.74}
\usepackage[pagebackref,breaklinks,colorlinks,allcolors=cvprblue]{hyperref}

\def\confName{CVPR}
\def\confYear{2027}

\title{LiteSTAR: Low-Rank Channel Modulation for Temporally Consistent CVD Video Compensation}

\author{
Peiyang Lin$^{1}$ \quad
Sichen Lin$^{1}$ \quad
Yipeng Liao$^{1}$ \quad
Liqun Lin$^{1,*}$ \quad
Xu Wang$^{2}$ \quad
Tiesong Zhao$^{1}$\\
$^{1}$Fuzhou University \quad
$^{2}$Shenzhen University\\
{\small
\texttt{112301516@fzu.edu.cn} (P. Lin),
\texttt{2601110040@fzu.edu.cn} (S. Lin)}\\
{\small
\texttt{fzu\_lyp@163.com} (Y. Liao),
\texttt{lin\_liqun@fzu.edu.cn} (L. Lin)}\\
{\small
\texttt{wangxu@szu.edu.cn} (X. Wang),
\texttt{t.zhao@fzu.edu.cn} (T. Zhao)}\\
{\small $^{*}$Corresponding author}
}
\usepackage{multirow}
\usepackage{booktabs}
\usepackage{xcolor}
\usepackage{graphicx}
\begin{document}
\maketitle
 \begin{abstract}
Color Vision Deficiency (CVD) affects a large population worldwide,
yet practical recoloring must run continuously on-device. Frame-wise video recoloring produces color flickering, while existing methods rely on costly cross-frame alignment in pixel or feature space. We observe that color flickering correlates with variations in channel-wise responses, with color-sensitive variations concentrated in a low-rank subspace. Based on this observation, we propose Lightweight Spatio-Temporal Adaptive Recoloring (LiteSTAR), built around LiCaD, a plug-and-play  Lightweight Channel-Adaptive Decoder. LiCaD performs temporal filtering in a low-rank state space with learned process and observation uncertainties. It then modulates feature channels to achieve temporal consistency without explicit spatial alignment. LiteSTAR further employs adaptive patch routing to improve color discrimination for CVD users while avoiding unnecessary over-enhancement. We also introduce EgoCVDSet, the first egocentric video benchmark for CVD recoloring. Extensive quantitative and qualitative experiments demonstrate that LiteSTAR achieves superior performance with only 0.05M parameters. LiCaD further demonstrates strong generalization across different tasks and network architectures. The code, models, and dataset will be released after review.
\end{abstract}
  \section{Introduction}

\begin{figure}[t]
    \centering
    \includegraphics[width=\columnwidth]{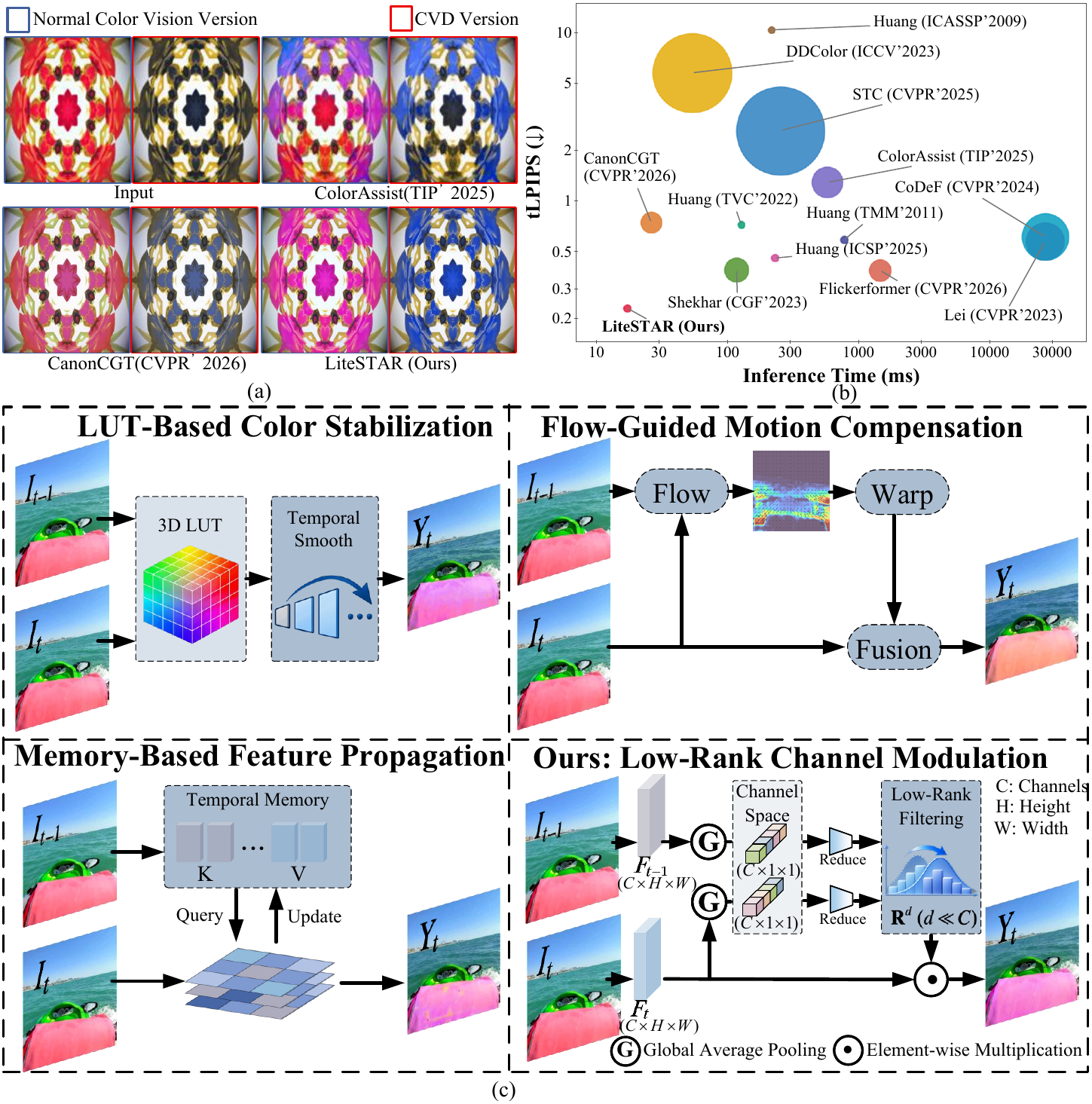}
    \caption{
    Workflow and performance comparison of our proposed
    LiteSTAR with existing CVD recoloring and temporal
    consistency methods.
(a) shows LiteSTAR achieves superior color contrast
for individuals with CVD while avoiding
color contrasts inconsistent with the original image. (b) shows LiteSTAR achieves superior
temporal consistency with minimal parameters
and fast inference.
The bubble size represents the number of model parameters. (c) illustrates our proposed temporal consistency
method, which achieves temporal consistency
through modulation in a low-rank channel space,
avoiding the high computational cost of
full-dimensional feature modulation.
    }
    \label{fig:framework}
\end{figure}

\label{sec:intro}
Color Vision Deficiency (CVD) affects approximately 8\% of men and 0.5\% of women worldwide~\cite{simunovic2010colour}, impairing color discrimination in daily life. However, most existing methods prioritize visual quality for normal vision while overlooking the needs of CVD users.

In real-world applications, uploading daily visual scenes to
the cloud may raise privacy concerns for individuals with CVD.
Therefore, there is a growing need for real-time video recoloring
that runs continuously on mobile devices.
However, existing CVD recoloring methods face three major challenges.
\textbf{First}, existing methods often incur high computational costs. Traditional methods rely on clustering or iterative optimization. Deep learning-based methods employ computationally expensive attention mechanisms or iterative diffusion models~\cite{huang2009recolorization,huang2022naturalness,lin2025colorassist}. These methods introduce substantial overhead, limiting real-time inference on resource-constrained devices.
\textbf{Second}, existing CVD recoloring methods often over-enhance color contrast to improve discriminability~\cite{lin2025colorassist,gillooly2025vit}. This may introduce artificial color differences and distort natural color relationships, as shown in Fig.~\ref{fig:framework}(a).
\textbf{Third}, as shown in Fig.~\ref{fig:framework}(c),
existing methods enforce temporal consistency
through optical flow alignment, cross-frame spatial
attention, or global color lookup tables
~\cite{liang2024flowvid,shin2025lut,yang2024colormnet}.
However, these methods may incur high computational
costs, suffer from inaccurate alignment or attention
matching, or introduce color distortion due to
over-smoothed mappings.

To address these limitations, we investigate the relationship between color flickering and channel modulation. We observe that color flickering in recolored videos is closely correlated with variations in channel modulation weights between adjacent frames. Based on this observation, we propose a plug-and-play Lightweight Channel-Adaptive Decoder (LiCaD). As shown in Fig.~\ref{fig:framework}(c), LiCaD avoids explicit spatial alignment by performing temporal state filtering and channel-wise modulation in a low-rank channel space. As shown in Fig.~\ref{fig:framework}(b), LiCaD achieves temporal consistency with minimal parameters and fast inference. Moreover, LiCaD can be integrated into diverse architectures and tasks, demonstrating its broad applicability.
Furthermore, we propose Lightweight Spatio-Temporal Adaptive Recoloring (LiteSTAR). As shown in Fig.~\ref{fig:framework}(a), LiteSTAR effectively enhances color contrast for CVD users while avoiding over-enhancement that introduces artificial color differences. In addition, we introduce EgoCVDSet, the first egocentric video dataset for CVD recoloring, providing a realistic benchmark for practical applications.
The main contributions are summarized as follows:

\begin{itemize}
    \item 
We propose LiCaD, a plug-and-play module that captures residual feature variations not explained by color statistics. It modulates channel weights in a low-rank
color-sensitive subspace to improve temporal
color consistency across video frames. LiCaD generalizes well across diverse network architectures and visual tasks.

    \item 
We propose LiteSTAR, which employs adaptive patch
routing to focus computation on color-confusable regions.
It processes low-resolution global images and local
patches through a shared network, reducing parameter
redundancy and providing implicit structural regularization.

    \item We introduce EgoCVDSet, the first high-resolution egocentric
video dataset for individuals with CVD.
It covers diverse real-world daily scenarios
and provides a benchmark for future research.
Experiments demonstrate that LiteSTAR achieves
state-of-the-art performance.
\end{itemize}
   \section{Related Work}

\subsection{Efficient Architectures for Image Enhancement}
Zhang et al.~\cite{zhang2025patchadapter} employ sequential global-to-local processing for low-resolution context modeling and high-resolution patch refinement. Luo et al.~\cite{luo2025diffshadow} use parallel U-Nets for global-guided local diffusion-based shadow removal.
Kong et al.~\cite{kong2021classsr}
routed patches to networks of different capacities
based on restoration difficulty.
Mao et al.~\cite{mao2024adarevd}
introduced adaptive patch exiting
to avoid redundant computation.
However, global-local architectures increase latency and parameters, while adaptive patch selection requires extra prediction modules.
In contrast, LiteSTAR shares parameters between global and local inputs and employs lightweight CVD-aware routing to reduce computational overhead.
\subsection{Recoloring for Color Vision Deficiency}

Huang et al.~\cite{huang2022naturalness} introduced compensation range constraints to balance color contrast and naturalness. Jiang et al.~\cite{jiang2023personalized} proposed CVD-GAN with disentangled latent representations for personalized CVD-friendly image generation. Lin et al.~\cite{lin2025colorassist} introduced ColorAssist, combining perception-guided feature extraction with a diffusion Transformer for CVD recoloring. For video recoloring, Huang et al.~\cite{huang2011temporal} employed motion estimation and spatiotemporal constraints for temporally consistent video recoloring for dichromats. Huang et al.~\cite{huang2025video} investigated perceptual discrepancy-based video recoloring for dichromacy. However, existing image-based methods often over-enhance already distinguishable regions, while traditional video recoloring methods may struggle to adapt to complex scenes. In contrast, LiteSTAR selectively enhances color-confusable regions and improves temporal consistency.
\subsection{Temporal Color Consistency in Videos}

Shin et al.~\cite{shin2025lut} proposed a video color
grading framework based on 3D LUTs.
Liu et al.~\cite{liu2024tcvc} employed bidirectional
feature propagation and self-regularization
for video colorization.
Yang et al.~\cite{yang2024colormnet} proposed
ColorMNet, which retrieves historical features
through key-value memory.
Zhou et al.~\cite{zhou2022alignment} introduced
iterative temporal alignment for video restoration.
Liang et al.~\cite{liang2024flowvid} proposed
FlowVid, which propagates editing information
using optical flow while handling inaccurate
motion estimates.
However, existing methods may suffer from limited adaptability, high overhead, or inaccurate temporal matching.
In contrast, LiCaD filters color statistics in a low-rank temporal space to modulate feature channels, avoiding optical flow, large memory banks, and direct color smoothing.

    \section{Method}
\subsection{Problem Formulation}
Existing image recoloring methods typically process entire images as $Y_t=\mathcal{F}(I_t)$, where $I_t$ and $Y_t$ denote the input and recolored images, and $\mathcal{F}$ is the recoloring network. However, color-confusable regions often occupy only a small portion of an image. Full-image processing increases computational costs and may cause unnatural color contrasts.

\begin{figure}[t]
    \centering
    \includegraphics[width=\linewidth]{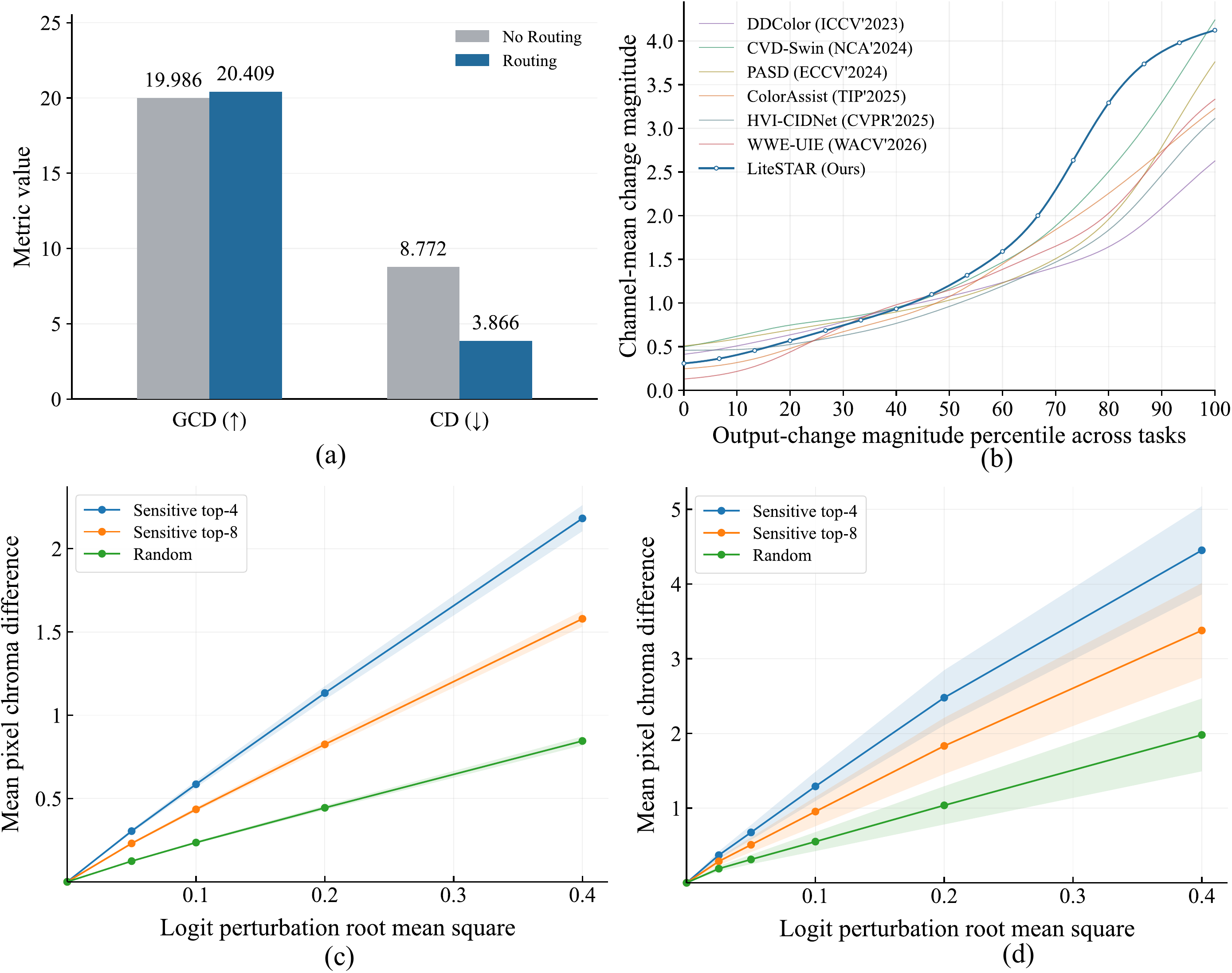}
    \caption{
        (a) Effect of adaptive routing on color contrast
        and preservation in low-CVD-confusion regions.
        (b) Cross-model relationship between temporal
        output variations and channel-wise feature
        drift~\cite{kang2023ddcolor,chen2024swin,
        yang2024pasd,lin2025colorassist,
        Yan_2025_CVPR,Cheng_2026_WACV}.
        (c) Relationship between channel-weight
        proportion perturbations and output color
        fluctuations in CVD recoloring.
        (d) Relationship between channel-weight
        proportion perturbations and output color
        fluctuations in underwater color correction. In both plots, Sensitive top-4 and top-8
denote perturbations along the four and eight
most color-sensitive directions, respectively,
while Random denotes perturbations along
randomly selected directions.
    }
    \label{fig:motivation_LiCaD1}
\end{figure}

To address these issues, we propose a global-local collaborative network with adaptive patch routing. It combines low-resolution global processing with selective local enhancement to improve efficiency and avoid over-enhancement, as shown in Fig.~\ref{fig:motivation_LiCaD1}(a). This process is formulated as $Y_t=\mathcal{F}(\operatorname{Downsample}(I_t),\operatorname{Route}(I_t^{\mathrm{patch}}))$, where $\operatorname{Downsample}(\cdot)$ reduces image resolution, $I_t^{\mathrm{patch}}$ denotes local patches, and $\operatorname{Route}(\cdot)$ selects color-confusable patches for enhancement.
\begin{figure*}[t]
    \centering
    \includegraphics[width=\linewidth]{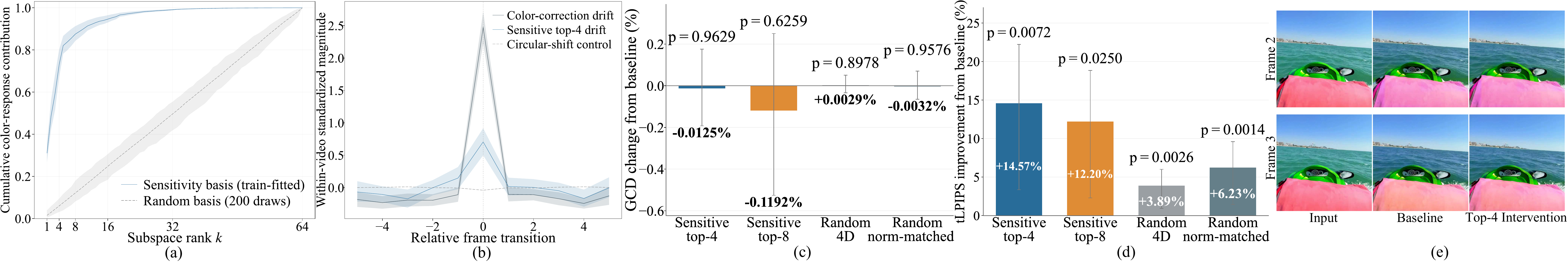}
    \caption{
(a) Generalization of color-sensitive
subspaces to unseen videos.
(b) Temporal correspondence between
sensitive-subspace drift and
color-correction changes.
(c) Effect of suppressing sensitive-subspace
drift on GCD.
(d) Effect of suppressing sensitive-subspace
drift on tLPIPS.
(e) Qualitative effects of
sensitive-subspace intervention. Random 4D uses a random four-dimensional subspace, while Random norm-matched matches the perturbation magnitude of Sensitive top-4.
    }
    \label{fig:motivation_LiCaD}
\end{figure*}

\subsection{From Image to Video Recoloring}

Existing methods typically require spatial alignment
of images or features between adjacent frames
to mitigate color flickering when extending
image-based recoloring models to video processing
~\cite{liu2024tcvc,zhou2022alignment}.
This process can be formulated as follows:
\begin{equation}
\begin{aligned}
Z_{t-1} &= \mathcal{E}(I_{t-1}), \quad
Z_t = \mathcal{E}(I_t),\\
Y_t &= \mathcal{D}\left(
Z_t, \operatorname{Warp}(Z_{t-1})
\right),
\end{aligned}
\label{eq:temporal_alignment}
\end{equation}
where $I_t$ and $I_{t-1}$ denote the current
and previous frames, respectively.
$\mathcal{E}$ denotes a feature extractor,
while $Z_t$ and $Z_{t-1}$ are the extracted features.
$\operatorname{Warp}(\cdot)$ denotes temporal
alignment, and $\mathcal{D}(\cdot)$ is a decoder. 
However, these methods often require
computationally expensive explicit spatial
alignment or cross-frame attention matching,
resulting in high computational overhead.

To address these issues, we revisit
the color flickering problem when extending
image-based models to video processing.
We investigate its relationship with
variations in channel-wise feature responses across adjacent frames.

\label{sec:formatting}
\noindent\textbf{\textit{
Verification A: Correlation between temporal color flickering
and channel-wise feature responses across adjacent frames.
}}
We investigate the correlation between temporal
output fluctuations and changes in channel-wise
feature responses across different models and tasks.
Experiments are conducted on EgoCVDSet,
SDSD Indoor~\cite{wang2021sdsd},
and UOT32~\cite{kezebou2019underwater},
covering CVD recoloring, grayscale image
colorization, style transfer, underwater
color correction, and low-light enhancement.
As shown in Fig.~\ref{fig:motivation_LiCaD1}(b),
variations in channel mean responses are
temporally correlated with abrupt changes
in model outputs across these tasks. 
Furthermore, as shown in
Fig.~\ref{fig:motivation_LiCaD1}(c) and (d),
we introduce controlled perturbations
to channel attention proportions in
CVD recoloring and underwater color correction tasks.
The results demonstrate a causal effect
of channel attention variations on
output color fluctuations. 
How do variations in the model's attention
to different feature channels
lead to color flickering?

\noindent\textbf{\textit{
Analysis B: Background-induced channel attention variations and color flickering.}} 
Temporal color flickering is often associated
with changes in global feature distributions
caused by newly appearing objects or
illumination variations.
Since different channels encode distinct
visual information, background or illumination
changes across adjacent frames may alter
the model's attention to different channels.
To illustrate this effect, we assume that
the features of the same object remain
stable across adjacent frames, denoted by
$S^c(p)$, where $p$ and $c$ represent
the spatial position and channel index,
respectively.
Let $B_t^c$ denote the modulation weight
of channel $c$ at frame $t$.
The modulated features of adjacent frames
can be expressed as

\begin{equation}
\begin{aligned}
F_{t-1}^c(p) &= B_{t-1}^c S^c(p),\\
F_t^c(p) &= B_t^c S^c(p).
\end{aligned}
\label{eq:adjacent_features}
\end{equation}

Although the object feature $S^c(p)$ remains unchanged, variations in channel responses can alter the output feature responses.
By subtracting the features of adjacent
frames and aggregating across channels,
we obtain
\begin{equation}
\Delta F_t(p) =
\sum_{c=1}^{C}
\Delta B_t^c S^c(p),
\label{eq:channel_variation}
\end{equation}
where $\Delta B_t^c = B_t^c-B_{t-1}^c$
denotes the change in channel modulation
responses.
This indicates that changes in channel
responses can alter the contributions of
different feature channels.
After color decoding, these differences
may further result in visible color flickering.

However, some channel variations provide
useful contextual information and enhance
feature representations.
Indiscriminately suppressing these variations
may degrade single-image performance.
This raises a key question: how can we
selectively suppress unstable channel variations
while preserving useful feature information?

\noindent\textbf{\textit{
Validation C: Verifying the low-rank structure of
channels sensitive to temporal color flickering.}}
We identify a low-rank, color-sensitive subspace via principal component analysis on the EgoCVD training set and evaluate its contribution on the unseen test set.
As shown in Fig.~\ref{fig:motivation_LiCaD}(a), most variations in color-sensitive responses are concentrated in a low-rank subspace. Fig.~\ref{fig:motivation_LiCaD}(b) further shows that temporal variations along these sensitive directions correlate with abrupt color changes in recolored videos. Moreover, as shown in Fig.~\ref{fig:motivation_LiCaD1}(c) and (d), perturbations along these directions induce more pronounced color fluctuations. These observations raise a key question: How can we suppress unstable variations in this low-rank subspace while preserving useful feature representations?

\noindent\textbf{\textit{
Analysis D: Low-rank decomposition and
selective temporal modulation.}} Motivated by this observation,
we decompose the channel attention
weight representation as follows:
\begin{equation}
\begin{aligned}
\mathbf{B}_t
&=
\underbrace{U_d U_d^\top \mathbf{B}_t}
_{\mathbf{B}_t^{\mathrm{sen}}}
+
\underbrace{(I-U_d U_d^\top)\mathbf{B}_t}
_{\mathbf{B}_t^{\mathrm{ins}}}.
\end{aligned}
\label{eq:lowrank_decomposition}
\end{equation}
where $U_d\in\mathbb{R}^{c\times d}$ denotes
an orthonormal basis of the color-sensitive
subspace, with $d\ll c$.
$\mathbf{B}_t^{\mathrm{sen}}$ and
$\mathbf{B}_t^{\mathrm{ins}}$ denote
the color-sensitive and insensitive
components.
This decomposition motivates temporal modeling
in the low-rank sensitive space
rather than across all feature dimensions.
Such selective modulation suppresses
color fluctuations while preserving
current-frame representations.
We learn a projection matrix to extract
$\mathbf{B}_t^{\mathrm{sen}}$.
Our goal is to stabilize color-sensitive directions through selective temporal modulation while preserving useful feature representations.

\begin{equation}
a_t^* =
\arg\min_{a_t}
\left\|
a_t \odot
\mathbf{B}_t^{\mathrm{sen}}
-
\mathbf{B}_{t-1}^{\mathrm{sen}}
\right\|^2.
\label{eq:sensitive_consistency}
\end{equation}
where $a_t$  denotes the channel-wise modulation weights.

\begin{figure*}[t]
    \centering
    \includegraphics[width=0.9\textwidth]{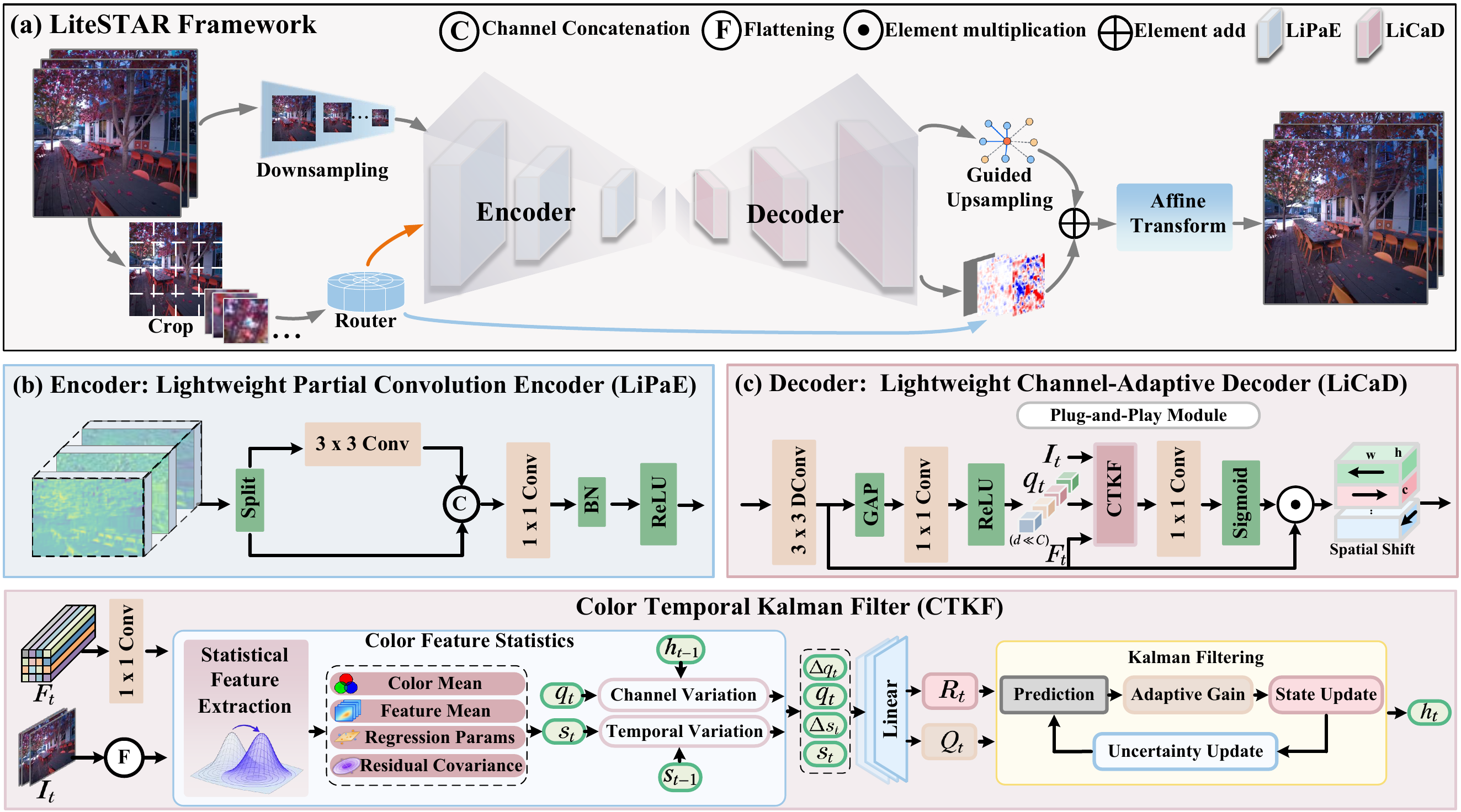}
\caption{
Overview of the proposed LiteSTAR framework.
(a) Overall architecture of LiteSTAR, featuring
global-local collaborative processing and adaptive patch routing.
(b) Structure of LiPaE used to construct the encoder.
(c) Structure of the LiCaD used to construct the decoder.
LiCaD incorporates a Color Temporal Kalman Filter (CTKF) through low-rank
channel state filtering and modulation.
}
    \label{fig:model}
\end{figure*}

\noindent\textbf{\textit{
Experiment E: Effectiveness of low-rank channel
modulation for temporal color consistency.}}
To preliminarily validate the effectiveness
of low-rank channel modulation, we transfer
the spatially averaged activations of selected
sensitive directions from the previous frame
to modulate current-frame features.
As shown in Figs.~\ref{fig:motivation_LiCaD}(c),
(d), and (e), even this simple strategy
improves temporal color consistency
without significantly degrading
single-image recoloring performance.
However, modulating more directions does not
consistently improve performance, suggesting that excessive historical
information may interfere with current-frame
representations.

Therefore, we propose temporal channel modulation
in a low-rank color-sensitive subspace.
Unlike conventional methods that require
high-dimensional spatial feature alignment,
our method models temporal consistency
in a compact channel space.
This avoids explicit spatial alignment
and reduces computational overhead.
The overall process can be formulated as follows:
\begin{equation}
\begin{aligned}
Z_{t-1} &=
\mathcal{E}\left(
I_{t-1}
\right),
Z_t =
\mathcal{E}\left(
I_t
\right),\\
a_t&=\mathcal{K}\left(
\mathrm{GAP}(Z_t),\mathrm{GAP}(Z_{t-1})
\right),\\
Y_t &= \mathcal{D}\left(
Z_t \odot a_t
\right).
\end{aligned}
\label{eq:channel_modulation}
\end{equation}
where $\mathrm{GAP}(\cdot)$ denotes global average pooling,
while $\mathcal{K}(\cdot)$ performs temporal state filtering to generate channel modulation weights.
$\odot$ denotes element-wise multiplication.

\subsection{Overview of Our Work}

We propose LiteSTAR, as illustrated in Fig.~\ref{fig:model}. \textbf{First}, adaptive patch routing selects color-confusable regions, while a shared network processes global thumbnails and selected patches. \textbf{Second,} the Lightweight Partial Convolution Encoder (LiPaE) extracts multi-scale features by convolving only one-quarter of the channels. \textbf{Third}, LiCaD enhances temporal color consistency through low-rank state filtering and channel modulation. \textbf{Finally}, color-guided upsampling and local refinement produce full-resolution affine parameters for CVD-friendly recoloring.

\begin{table*}[t]
\centering
\caption{Quantitative comparisons on the FZU-CVDSet and Li's datasets. The best and second-best
results are highlighted in red and blue. }
\label{tab:comparison}

\setlength{\tabcolsep}{4pt}
\renewcommand{\arraystretch}{1.15}

\resizebox{0.75\textwidth}{!}{
\begin{tabular}{l|l|ccccccccc}

\toprule
\multirow{2}{*}{Dataset}
& \multirow{2}{*}{Criterion}
& Huang
& Huang
& DDColor
& PASD
& EnColor
& Gillooly
& ColorAssist
& CanonCGT
& \multirow{2}{*}{LiteSTAR} \\

& 
& (ICASSP'2009)
& (Vis Comput'2022)
& (ICCV'2023)
& (ECCV'2024)
& (AAAI'2024)
& (WACV'2025)
& (TIP'2025)
& (CVPR'2026)
& \\

\midrule


\multirow{4}{*}{FZU-CVDSet}
& GCD ($\uparrow$)
& 39.0891
& 39.2654
& 38.9131
& 27.9082
& 38.3619
& 39.7925
& \textcolor{red}{41.3779}
& 34.8334
& \textcolor{blue}{40.4543} \\

& CD ($\downarrow$)
& 10.2342
& 9.8093
& 11.9870
& 12.3368
& 12.2823
& \textcolor{blue}{7.3152}
& 7.6067
& 10.7134
& \textcolor{red}{2.2176} \\

& PSNR ($\uparrow$)
& 13.8507
& 16.3607
& 13.0412
& 13.3461
& 12.6040
& \textcolor{red}{24.0130}
& 18.5766
& 15.1378
& \textcolor{blue}{22.1290} \\

& SSIM ($\uparrow$)
& 0.7408
& 0.8148
& 0.6962
& 0.7565
& 0.4799
& \textcolor{blue}{0.9313}
& 0.8227
& 0.8272
& \textcolor{red}{0.9494} \\

\midrule


\multirow{4}{*}{Li's Dataset}
& GCD ($\uparrow$)
& 40.6633
& 39.9154
& 33.8204
& 30.2250
& 39.7978
& 39.7289
& \textcolor{red}{43.3546}
& 33.2660
& \textcolor{blue}{41.0254} \\

& CD ($\downarrow$)
& 10.8315
& 10.2327
& 12.0401
& 12.4174
& 12.6119
& \textcolor{blue}{7.7047}
& 8.3525
& 7.9269
& \textcolor{red}{2.8975} \\

& PSNR ($\uparrow$)
& 12.7672
& 16.6441
& 16.5148
& 13.3584
& 11.5310
& \textcolor{blue}{23.0209}
& 17.0671
& \textcolor{red}{24.3714}
& 21.1852 \\

& SSIM ($\uparrow$)
& 0.7081
& 0.8121
& 0.6859
& 0.7721
& 0.5836
& 0.9122
& 0.7508
& \textcolor{blue}{0.9134}
& \textcolor{red}{0.9306} \\

\bottomrule
\end{tabular}
}
\end{table*}

\subsection{LiCaD}
Based on the above analysis, we globally aggregate the current-frame features $F_t$ to obtain channel-wise responses. We replace PCA decomposition with learnable convolutions for better robustness across scenes, reducing the channel dimension to obtain a low-rank representation $q_t$. To distinguish channel response variations caused by perturbations along sensitive directions from those induced by input changes, we project and flatten $F_t$ into $U_t$ and spatially align and flatten $I_t$ into $G_t$. Their color-feature relationship is modeled using linear regression and residual covariance:
\begin{equation}
\begin{aligned}
A_t
&=
\Sigma_{\mathrm{UG},t}
\left(
\Sigma_{\mathrm{GG},t}
+
\varepsilon E_3
\right)^{-1},
\\
b_t
&=
\mu_{U,t}-A_t\mu_{G,t},
\\
Z_t
&=
U_t-A_tG_t-b_t\mathbf{1}^{\top},
\\
V_t
&=
\frac{Z_tZ_t^{\top}}{N-1}
+
\varepsilon E_d,
\end{aligned}
\end{equation}
where $U_t\in\mathbb{R}^{d\times N}$ and
$G_t\in\mathbb{R}^{3\times N}$ denote the
aligned feature and color matrices,
with $N$ spatial samples and $d=C/16$.
$A_t$ and $b_t$ are the regression matrix
and bias, while $Z_t$ and $V_t$ denote
the residual features and their covariance.
$\mu_{U,t}$ and $\mu_{G,t}$ are feature
and color means, and $\Sigma_{UG,t}$
and $\Sigma_{GG,t}$ are the cross-covariance
and color covariance, respectively.
$E_3$ and $E_d$ are identity matrices,
and $\varepsilon$ ensures numerical stability.

The color-conditioned descriptor is defined as:
\begin{equation}
s_t = \operatorname{Concat}\big[
\mu_{G,t}, \mu_{U,t},
\operatorname{vec}(A_t),
b_t, \operatorname{svec}(\log V_t)
\big].
\end{equation}
where $\operatorname{vec}$ and $\operatorname{svec}$
denote standard and symmetric vectorization,
respectively.

We then predict adaptive process and observation
uncertainties from temporal changes in channel
observations and color statistics:
\begin{equation}
\begin{aligned}
\Delta q_t &= q_t-h_{t-1}, \quad
\Delta s_t=s_t-s_{t-1},\\
[Q_t,R_t]
&=\operatorname{Linear}\left(
\operatorname{Concat}
[s_t,\Delta s_t,q_t,\Delta q_t]
\right).
\end{aligned}
\label{eq:adaptive_uncertainty}
\end{equation}
where $h_{t-1}$ is the previous filtered state,
and $Q_t,R_t\in\mathbb{R}^{d}$ are the predicted
uncertainties. We employ Kalman filtering~\cite{kalman1960new} as a tool to achieve temporal stability in the low-rank feature space, producing a temporally consistent state $h_t$. A convolution then projects $h_t$ back to the original channel dimension, yielding channel modulation weights $a_t$:
\begin{equation}
\widetilde{F}_t = F_t \odot a_t.
\end{equation}
\textcolor{red}{\textbf{More implementation details are provided in Section 1 of the supplementary material.}}
\begin{table*}[t]
\centering
\caption{
Quantitative comparison of recoloring and temporal consistency methods on EgoCVDSet.
The best and second-best results are highlighted in red and blue. All methods are evaluated on an AMD Ryzen 9 7945HX CPU
and an NVIDIA RTX 4070 Laptop GPU.
}
\label{tab:all_metrics_comparison}

\setlength{\tabcolsep}{3.5pt}
\renewcommand{\arraystretch}{1.15}

\resizebox{0.77\textwidth}{!}{
\begin{tabular}{l|l|ccccccccc}
\toprule
Category & Model
& WE $\downarrow$
& tLPIPS $\downarrow$
& GCD $\uparrow$
& CD $\downarrow$
& PSNR $\uparrow$
& SSIM $\uparrow$
& Time (ms) $\downarrow$
& GFLOPs $\downarrow$
& Params (M) $\downarrow$ \\
\midrule


\multirow{5}{*}{Single-image}
& Huang (ICASSP'2009)
& 3.7508 & 10.4019
& 29.2513 & 8.7520
& 19.5383 & 0.8684 & 216.6260
& -- & -- \\

& Huang (Vis Comput'2022)
& 0.6627 & 0.7203
& \textcolor{blue}{29.9056} & 7.3428
& 24.8282 & 0.9224 & 127.4422
& -- & -- \\

& DDColor (ICCV'2023)
& 2.2414 & 5.7785
& 27.8190 & 11.4355
& 17.8923 & 0.8226 & 53.8077
& 50.3144 & 55.0066 \\

& ColorAssist (TIP'2025)
& 1.3641 & 1.2892
& 28.8139 & 5.5014
& 24.5085 & 0.8027 & 579.8431
& 2459.5780 & 8.0123 \\

& CanonCGT (CVPR'2026)
& 0.8674 & 0.7422
& 26.0415 & 9.6788
& 23.6608 & 0.9126 & 26.2586
& 10.9430 & 3.6906 \\

\midrule


\multirow{5}{*}{Post-training}
& + Lei (CVPR'2023)
& 0.7982 & 0.5762
& 27.7247 & 3.9309
& 30.1800 & 0.9546
& +26578.9629 & +294.0582 & +12.2891 \\

& + Shekhar (CGF'2023)
& 0.7427 & 0.3908
& 28.4813 & 3.3874
& 32.3834 & 0.9678
& +105.7656 & +28.7888 & +4.8534 \\

& + CoDeF (CVPR'2024)
& 0.8080 & 0.6106
& 28.1634 & 3.7275
& 30.6181 & 0.9389
& +26760.3308 & +19.6629 & +18.9378 \\

& + STC (CVPR'2025)
& 1.6791 & 2.6196
& 27.8467 & 7.2359
& 23.0530 & 0.7876
& +242.1132 & +982.1504 & +67.9073 \\

& + Flickerformer (CVPR'2026)
& 0.8028 & 0.3866
& 28.1140 & 3.6444
& 31.4454 & 0.9638
& +1449.7035 & +261.2167 & +4.1551 \\

\midrule

\multirow{2}{*}{Video processing}
& Huang (TMM'2011)
& 0.7064
& 0.5873
& 28.0088
& 9.2395
& 20.4169
& 0.8576
& 777.2918
& -- & -- \\

& Huang (ICSP'2025)
& 0.5353
& 0.4576
& 29.2925
& 7.2773
& 22.2435
& 0.9209
& 229.7930
& -- & -- \\

\midrule


\multirow{2}{*}{Ours}
& LiteSTAR (w/o LiCaD)
& \textcolor{blue}{0.5062}
& \textcolor{blue}{0.3442}
& 28.5390
& \textcolor{blue}{3.0313}
& \textcolor{blue}{32.6740}
& \textcolor{red}{0.9686}
& \textcolor{red}{11.6613}
& \textcolor{red}{5.8722}
& \textcolor{red}{0.0498} \\

& LiteSTAR
& \textcolor{red}{0.4573}
& \textcolor{red}{0.2305}
& \textcolor{red}{30.0129}
& \textcolor{red}{2.4528}
& \textcolor{red}{32.7882}
& \textcolor{blue}{0.9685}
& \textcolor{blue}{17.1491}
& \textcolor{blue}{5.9559}
& \textcolor{blue}{0.0524} \\

\bottomrule
\end{tabular}
}
\vspace{1mm}
\end{table*}

\subsection{EgoCVDSet}
Due to the lack of egocentric video datasets tailored to CVD users, existing evaluations may not fully reflect real-world needs. To address this gap, we construct EgoCVDSet, an egocentric video dataset for CVD recoloring. \textcolor{red}{\textbf{Further details on dataset construction and statistics are provided in Section 2 of the supplementary material.}}

     \section{Experiments}

\begin{figure*}[t]
    \centering
    \includegraphics[width=\textwidth]{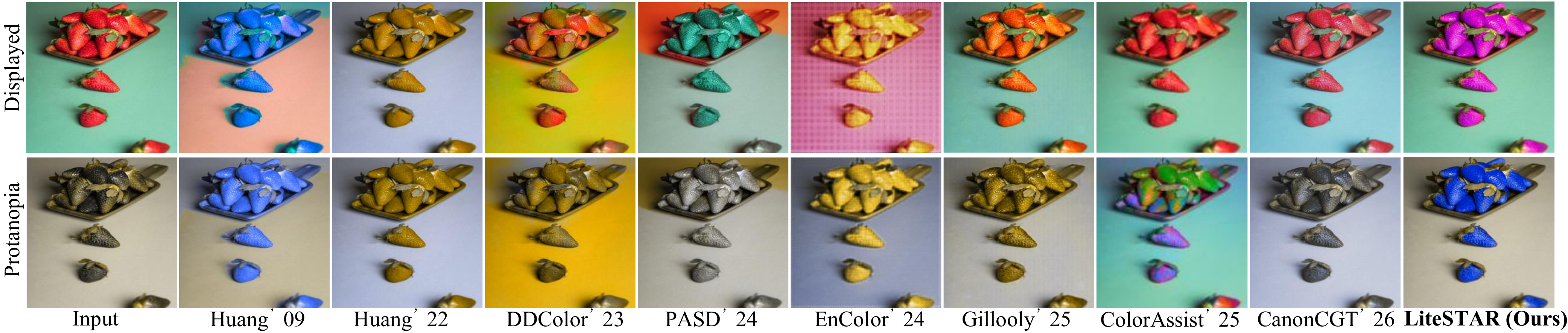}
    \caption{Qualitative comparison of CVD recoloring methods
on the FZU-CVDSet dataset. The first row shows the original input and recolored images, while the second row
shows their corresponding protanopia simulations.}
    \label{fig:contrast1}
\end{figure*}

\subsection{Data and Evaluation Metrics}
\noindent\textbf{\textit{1) Datasets:}}
First, image recoloring models are trained on the FZU-CVDSet image dataset and evaluated on FZU-CVDSet~\cite{lin2025colorassist} and Li's dataset~\cite{li2020cvd}. Second, LiCaD and other temporal post-training methods are trained on the EgoCVDSet video dataset with their backbones frozen.

\noindent\textbf{\textit{2) Evaluation Metrics:}} 
To evaluate the effectiveness of LiteSTAR for
individuals with CVD, we adopt Global Chromatic
Diversity (GCD)~\cite{ma2009color} and Chromatic
Difference (CD)~\cite{zhu2021survey}.
Following~\cite{seo2025cudnet}, we adopt Peak Signal-to-Noise Ratio (PSNR) and Structural Similarity Index (SSIM) to evaluate pixel-level fidelity and structural similarity between the original and recolored images.
Furthermore, to evaluate temporal consistency
in video recoloring, we adopt Warping Error (WE)~\cite{lai2018temporal}
and Temporal Learned Perceptual Image Patch Similarity (tLPIPS)~\cite{chu2020temporal}. WE and tLPIPS are scaled by $10^2$. \textcolor{red}{\textbf{We also conduct a subjective evaluation, as detailed in Section 3 of the supplementary material.}}

\noindent\textbf{\textit{3) Implementation Details:}}
First, the image recoloring network is trained for
370 epochs on $256\times256$ images,
with global thumbnails and local patches at
$64\times64$ resolution.
Then, we freeze the backbone and temporally
train LiCaD for 100 epochs on four-frame
$512\times512$ clips, using a learning rate
of $10^{-4}$ for both stages.

\subsection{State-of-the-Art Comparison }
To evaluate LiteSTAR, we compare it with existing
CVD recoloring methods~\cite{huang2009recolorization,
huang2022naturalness,goyal2024encolor,
gillooly2025vit,lin2025colorassist}, Video processing~\cite{huang2011temporal,huang2025video}, general image color processing methods~
\cite{kang2023ddcolor,yang2024pasd,ko2026canoncgt} and video post-training methods applied to the
single-image LiteSTAR backbone~
\cite{lei2023deflicker,shekhar2023temporal,
ouyang2024codef,wang2025efficient,qu2026flickerformer}. LiteSTAR (w/o LiCaD) disables temporal channel modulation in LiCaD while retaining its original decoding functionality.

\noindent\textbf{\textit{1) Quantitative Evaluation:}}
Table~\ref{tab:comparison} shows that LiteSTAR
improves color discrimination while preserving
image structure and naturalness.
Gillooly~\cite{gillooly2025vit} and
CanonCGT~\cite{ko2026canoncgt} achieve higher PSNR
on Li's Dataset through conservative adjustments,
but provide limited contrast enhancement.
Although ColorAssist~\cite{lin2025colorassist}
achieves higher GCD, it introduces excessive
color contrast.
In contrast, LiteSTAR better balances
color discrimination and visual fidelity.
Table~\ref{tab:all_metrics_comparison} shows that
LiCaD achieves strong temporal color consistency
with minimal computational overhead.

\noindent\textbf{\textit{2) Qualitative Evaluation:}}
Fig.~\ref{fig:contrast1} shows that Huang~\cite{huang2009recolorization}, DDColor~\cite{kang2023ddcolor}, PASD~\cite{yang2024pasd}, and ColorAssist~\cite{lin2025colorassist}
introduce artificial color contrasts
that are absent from the original images.
In contrast, Huang~\cite{huang2022naturalness}, Gillooly~\cite{gillooly2025vit}, and CanonCGT~\cite{ko2026canoncgt}
largely preserve the original colors but provide
limited contrast enhancement for individuals with CVD.
Moreover, EnColor~\cite{goyal2024encolor} produces blurry results
with noticeable loss of image details.
Our LiteSTAR effectively enhances color
discrimination for individuals with CVD
while preserving original color relationships
and image details, avoiding over-enhancement.

\begin{figure*}[t]
    \centering
    \includegraphics[
        width=0.9\textwidth,
        height=0.66\textheight
    ]{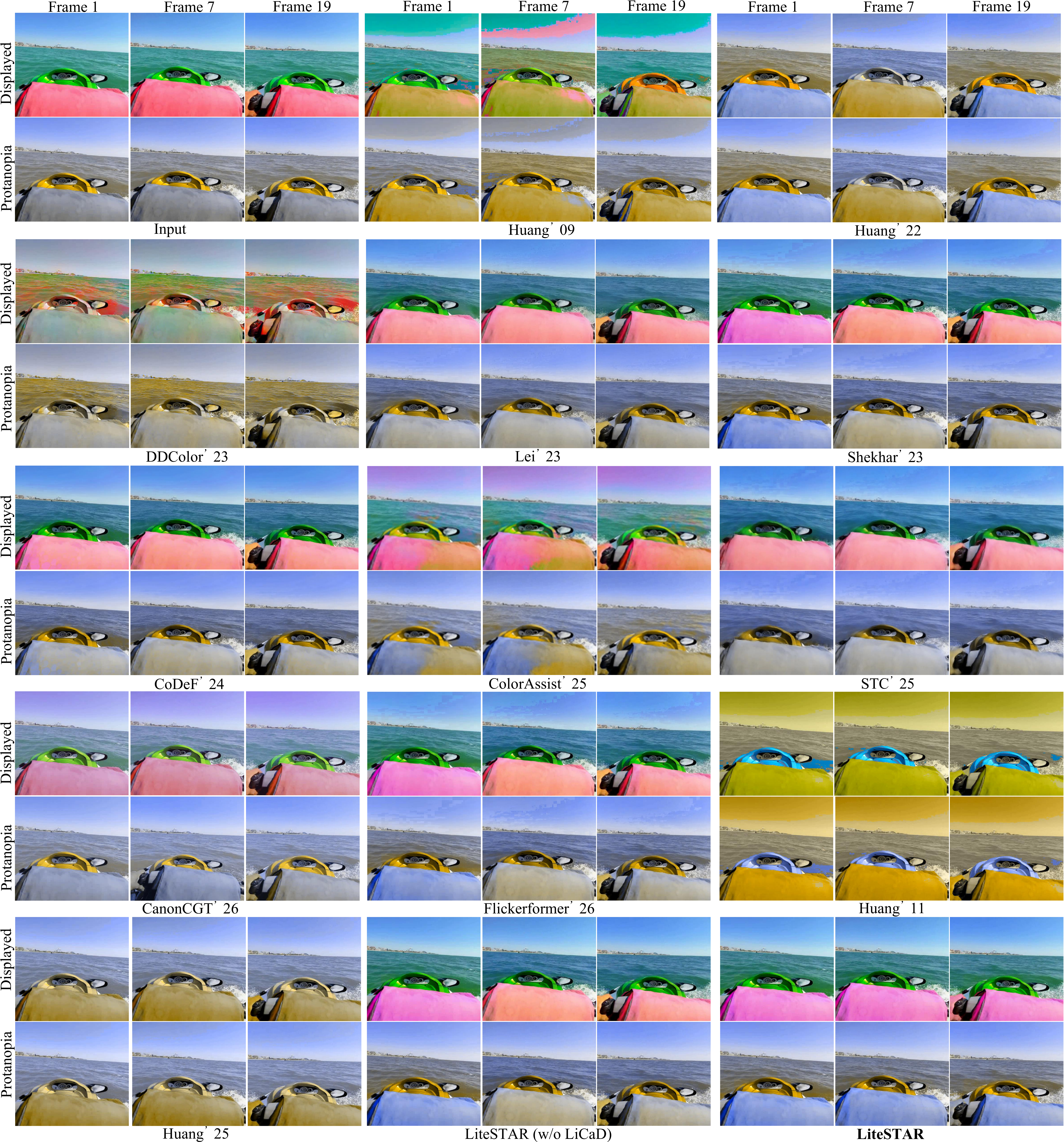}
    \caption{Qualitative comparison of video recoloring methods
    on the EgoCVDSet dataset.
For each method, the upper and lower rows show the displayed images and protanopia simulations, respectively.}
    \label{fig:contrast2}
\end{figure*}

Fig.~\ref{fig:contrast2} shows that single-image recoloring methods
suffer from noticeable temporal flickering
and color distortion.
Shekhar~\cite{shekhar2023temporal} and Flickerformer~\cite{qu2026flickerformer} still struggle to
eliminate large-scale color fluctuations.
Although Lei~\cite{lei2023deflicker}, CoDeF~\cite{ouyang2024codef}, and STC~\cite{wang2025efficient}
achieve more stable global color distributions,
they may weaken color contrast under CVD
perception and introduce local color inconsistencies. Meanwhile, existing traditional CVD video recoloring methods still struggle to effectively enhance color contrast and maintain temporal consistency.
In contrast, LiCaD suppresses temporal color
fluctuations while preserving color contrast,
achieving improved temporal consistency.

\subsubsection{LiCaD Generalization Across Architectures}

To evaluate the plug-and-play
capability of LiCaD, we integrate it into
Swin Transformer-based Chen~\cite{chen2024swin},
diffusion Transformer-based ColorAssist~\cite{lin2025colorassist},
Stable Diffusion-based PASD~\cite{yang2024pasd},
and 3D LUT-based CanonCGT~\cite{ko2026canoncgt}.
Table~\ref{tab:LiCaD_comparison} shows that LiCaD improves temporal consistency while largely preserving single-image performance.

\begin{table}[t]
\centering
\caption{
Quantitative comparisons before and after
integrating LiCaD on EgoCVDSet. 
}
\label{tab:LiCaD_comparison}

\setlength{\tabcolsep}{3pt}
\renewcommand{\arraystretch}{1.12}

\resizebox{\columnwidth}{!}{
\begin{tabular}{lcccccc}
\toprule
Method
& WE $\downarrow$
& tLPIPS  $\downarrow$
& CD $\downarrow$
& GCD $\uparrow$
& PSNR $\uparrow$
& SSIM $\uparrow$ \\
\midrule

Chen (NCA'2024)
& 0.8733
& 0.4168
& 6.2221
& \textbf{31.4076}
& 23.2296
& 0.8858 \\

+ LiCaD
& \textbf{0.7829}
& \textbf{0.3772}
& \textbf{5.5533}
& 28.9318
& \textbf{26.2909}
& \textbf{0.9053} \\

\midrule

PASD (ECCV'2024)
& 6.9806
& 12.1200
& 12.6582
& \textbf{30.3143}
& 15.5265
& 0.7315 \\

+ LiCaD
& \textbf{1.3002}
& \textbf{0.9179}
& \textbf{12.4840}
& 26.1534
& \textbf{19.9575}
& \textbf{0.8481} \\

\midrule

ColorAssist (TIP'2025)
& 1.3641
& 1.2892
& \textbf{5.5014}
& 28.8139
& \textbf{24.5085}
& \textbf{0.8027} \\

+ LiCaD
& \textbf{1.0565}
& \textbf{0.9337}
& 7.1410
& \textbf{29.2110}
& 24.2125
& 0.7904 \\

\midrule

CanonCGT (CVPR'2026)
& 0.8674
& 0.7422
& 9.6788
& 26.0415
& 23.6608
& 0.9126 \\

+ LiCaD
& \textbf{0.4562}
& \textbf{0.2429}
& \textbf{3.6456}
& \textbf{27.3792}
& \textbf{34.2242}
& \textbf{0.9741} \\

\bottomrule
\end{tabular}
}
\end{table}

\begin{table}[t]
\centering
\caption{
Quantitative comparison of different temporal
consistency methods across CIDNet~\cite{Yan_2025_CVPR} and WWE-UIE~\cite{Cheng_2026_WACV}. The best and second-best
results are highlighted in red and blue.
}
\label{tab:cross_task_temporal}

\setlength{\tabcolsep}{3pt}
\renewcommand{\arraystretch}{1.15}
\captionsetup{justification=centering}

\begin{subtable}[t]{\columnwidth}
\centering
\caption{Results on CIDNet.}
\label{tab:cidnet_temporal_comparison}

\resizebox{\linewidth}{!}{
\begin{tabular}{l|ccccccc}
\toprule
\multirow{2}{*}{Criterion}
& \multirow{2}{*}{CIDNet}
& $+$Lei
& $+$Shekhar
& $+$CoDeF
& $+$STC
& $+$Flickerformer
& \multirow{2}{*}{$+$LiCaD} \\

&
&
{\scriptsize (CVPR'2023)}
& {\scriptsize (CGF'2023)}
& {\scriptsize (CVPR'2024)}
& {\scriptsize (CVPR'2025)}
& {\scriptsize (CVPR'2026)}
& \\
\midrule

WE  $\downarrow$
& 1.8773
& 1.0867
& 1.0121
& \textcolor{blue}{0.9727}
& 1.2141
& 1.6736
& \textcolor{red}{0.5908} \\

tLPIPS  $\downarrow$
& 8.0685
& 5.3901
& 3.8428
& \textcolor{red}{1.1997}
& 1.8656
& 7.3686
& \textcolor{blue}{1.2710} \\

PSNR $\uparrow$
& 16.1791
& \textcolor{blue}{16.2849}
& 16.2796
& \textcolor{red}{16.3337}
& 16.0869
& 15.9573
& 16.2179 \\

SSIM $\uparrow$
& 0.6775
& 0.6955
& 0.7038
& \textcolor{blue}{0.7245}
& 0.7154
& 0.6742
& \textcolor{red}{0.7526} \\

\bottomrule
\end{tabular}
}

\end{subtable}

\vspace{3mm}

\begin{subtable}[t]{\columnwidth}
\centering
\caption{Results on WWE-UIE.}
\label{tab:wwe_uie_temporal}

\resizebox{\linewidth}{!}{
\begin{tabular}{l|ccccccc}
\toprule
\multirow{2}{*}{Criterion}
& \multirow{2}{*}{WWE-UIE}
&$+$Lei
& $+$Shekhar
& $+$CoDeF
& $+$STC
& $+$Flickerformer
& \multirow{2}{*}{$+$LiCaD} \\

&
&
{\scriptsize (CVPR'2023)}
& {\scriptsize (CGF'2023)}
& {\scriptsize (CVPR'2024)}
& {\scriptsize (CVPR'2025)}
& {\scriptsize (CVPR'2026)}
& \\
\midrule

WE $\downarrow$
& 0.4135
& 0.7423
& \textcolor{blue}{0.3917}
& 0.8260
& 0.9882
& 0.4388
& \textcolor{red}{0.3558} \\

tLPIPS $\downarrow$
& 1.5038
& \textcolor{blue}{0.5623}
& 0.6566
& 1.7715
& 1.3191
& 1.1355
& \textcolor{red}{0.5438} \\

UIQM $\uparrow$
& 3.3592
& 3.3452
& 3.3886
& 3.2584
& 3.2636
& \textcolor{red}{3.4377}
& \textcolor{blue}{3.4333} \\

UCIQE $\uparrow$
& 0.3469
& 0.3228
& \textcolor{blue}{0.3472}
& 0.3351
& \textcolor{red}{0.3555}
& 0.3437
& 0.3222 \\

\bottomrule
\end{tabular}
}

\end{subtable}

\end{table}

\subsubsection{LiCaD Generalization Across Tasks}
Tables~\ref{tab:cidnet_temporal_comparison} and~\ref{tab:wwe_uie_temporal} show that LiCaD improves temporal consistency in low-light enhancement and underwater color correction while maintaining competitive image quality, demonstrating its generalization capability.

\textcolor{red}{\textbf{Detailed experimental settings and analysis are provided in Section 5 of the supplementary material.}}

\begin{table}[!t]
\centering
\caption{Ablation study of key design choices. }
\label{tab:ablation}

\setlength{\tabcolsep}{3pt}
\renewcommand{\arraystretch}{1.05}

\resizebox{0.7\columnwidth}{!}{
\begin{tabular}{l|cccc}
\toprule
Method
& CD $\downarrow$
& GCD $\uparrow$
& PSNR $\uparrow$
& SSIM $\uparrow$ \\
\midrule

High-Freq. Modulation
& 2.1930 & 38.9568 & 22.7016 & 0.9500 \\

Full-Image Input
& 6.1614 & 39.5347 & 23.0206 & 0.9387 \\

Unshared Networks
& \textbf{2.1103} & 37.9855
& \textbf{24.4364} & \textbf{0.9619} \\

w/o Patch Routing
& 6.0967 & \textbf{41.0614}
& 21.2588 & 0.9353 \\

\midrule

\textbf{Ours}
& 2.2176 & 40.4543 & 22.1290 & 0.9494 \\

\bottomrule
\end{tabular}
}
\vspace{1mm}
\end{table}

\subsection{Ablation Study}
\noindent\textbf{\textit{1) Ablation Study on the Key Design Choices of LiteSTAR:}} 
As shown in Table~\ref{tab:ablation}, high-frequency modulation, full-resolution processing, and unshared global-local networks limit the model's representation capacity, leading to lower GCD. Removing patch routing increases GCD but causes significant color over-enhancement.

\begin{table}[t]
\centering
\caption{Quantitative comparison of different temporal
consistency strategies and our methods. }
\label{tab:temporal_strategies_comparison}

\setlength{\tabcolsep}{3pt}
\renewcommand{\arraystretch}{1.15}

\resizebox{\columnwidth}{!}{
\begin{tabular}{l|cccccc}
\toprule
Method
& WE$\downarrow$
& tLPIPS$\downarrow$
& CD $\downarrow$
& GCD $\uparrow$
& PSNR $\uparrow$
& SSIM $\uparrow$ \\
\midrule

LUT-Based
& 0.4906
& 0.3220
& 3.6395
& 28.5739
& 28.3486
& 0.9687 \\

Memory-Based
& 0.5104
& 0.5029
& 2.5229
& 28.5212
& 29.7840
& 0.9665 \\

Flow-Guided
& 0.4850
& 0.2673
& 2.5476
& 28.3496
& 30.3027
& \textbf{0.9717} \\

\midrule

LiteSTAR (w/o LiCaD)
& 0.5062
& 0.3442
& 3.0313
& 28.5390
& 32.6740
& 0.9686 \\

LiteSTAR
& \textbf{0.4573}
& \textbf{0.2305}
& \textbf{2.4528}
& \textbf{30.0129}
& \textbf{32.7882}
& 0.9685 \\

\bottomrule
\end{tabular}
}
\end{table}

\noindent\textbf{\textit{2) Ablation on Temporal Consistency Strategies:}} 
Table~\ref{tab:temporal_strategies_comparison}
shows that our method outperforms LUT-based,
memory-based, and flow-guided methods
in temporal consistency.

\textcolor{red}{\textbf{Additional ablation studies are provided in Section 4 of the supplementary material.}}
      \section{Conclusion}

In this paper, we propose LiteSTAR, incorporating a plug-and-play LiCaD. Unlike spatial alignment-based methods, LiCaD suppresses temporal variations along color-sensitive feature directions in a low-rank subspace. It achieves strong temporal consistency with minimal parameter overhead and generalizes well across diverse model architectures and visual tasks. LiteSTAR employs adaptive patch routing to focus computation on color-confusable regions while avoiding unnecessary over-enhancement. We also introduce EgoCVDSet, the first egocentric video dataset specifically designed for CVD recoloring, providing a benchmark for future research.

{
    \small
    \bibliographystyle{ieeenat_fullname}
    \bibliography{main}
}


\end{document}